\documentclass[11pt,letterpaper,pdftex]{article}
\usepackage{cogsys}
\usepackage[most]{tcolorbox}
\usepackage{graphicx}
\usepackage[T1]{fontenc}
\usepackage{times}
\usepackage{booktabs}
\usepackage{multirow}

\usepackage{natbib}
\usepackage{enumitem}
\ShortHeadings{Basic Categories in Vision Language Models}
              {H.\ Sawyer, J.\ Roberts, and K.\ Moore}

\begin{document}

\title{Basic Categories in Vision Language Models: \\  Expert Prompting Doesn't Grant Expertise}
 
\author{Hunter Sawyer}{htsawyer42@tntech.edu}
\address{Computer Science, Tennessee Tech University, Cookeville, TN 38505 USA}
\author{Jesse Roberts}{jtroberts@tntech.edu}
\address{Computer Science, Tennessee Tech University, Cookeville, TN 38505 USA}
\author{Kyle Moore}{kyle.a.moore@Vanderbilt.Edu}
\address{Computer Science, Vanderbilt University, Nashville, TN 37235 USA}
 
\begin{abstract}The field of psychology has long recognized a basic level of categorization that humans use when labeling visual stimuli, a term coined by Rosch in 1976. This level of categorization has been found to be used most frequently, to have higher information density, and to aid in visual language tasks with priming in humans. Here, we investigate basic-level categorization in two recently released, open-source vision-language models (VLMs). This paper demonstrates that Llama 3.2 Vision Instruct (11B) and Molmo 7B-D both prefer basic-level categorization consistent with human behavior. Moreover, the models’ preferences are consistent with nuanced human behaviors like the biological versus non-biological basic level effects and the well-established expert basic level shift, further suggesting that VLMs acquire complex cognitive categorization behaviors from the human data on which they are trained. We also find our expert prompting methods demonstrate lower accuracy then our non-expert prompting methods, contradicting popular thought regarding the use of expertise prompting methods.
\end{abstract}

\section{Introduction}
\tcbset{
    mymsg/.style={
        enhanced,
        colframe=white, frame hidden,
        boxrule=0pt,
        arc=18pt,
        outer arc=18pt,
        boxsep=2pt, left=8pt, right=8pt, top=6pt,
        colback=#1!25,
        fuzzy shadow={1mm}{-0.7mm}{0mm}{0.1mm}{black!40!white},
        width=0.8\linewidth,
        before skip=-0.5mm, 
    },
    expertstyle/.style={mymsg=green, enlarge right by=10mm},
    llmstyle/.style={mymsg=blue, enlarge left by=10mm}
}

The growing field of natural language processing (NLP) has received significant public attention in recent years due to the rapid increase in availability of large language models (LLMs). Often, these models use variations of the transformer architecture, which has dramatically improved both the accuracy and training speed of NLP tasks \citep{vaswani_attention_2023}. 

NLP is fundamentally linked to humans, originating from the generation of natural language, and it can serve as a medium for expressing complex human cognitive processes. This fundamental property of NLP has several implications for the relation between language models and humans, with this paper specifically focusing on the psychological relationship.

The core interest of this study is the notion that if a model is trained to generate text based on human-produced language, it may be learning underlying human cognitive or psychological behaviors. These may even include abstract behaviors in humans such as logical reasoning or object categorization, with object categorization being the behavior of focus in this work.

Here, we examine a psychological phenomenon known as ~``basic-level'' categorization \citep{rosch_basic_1976} which has several beneficial properties. This phenomenon and its foundation in the cognitive literature, will be examined more in the background section. Briefly, it refers to the most commonly used level of categorization that has utility in communication because of its elevated level of category differentiation and information density. The fundamental question being addressed in this paper is whether basic-level categorization behaviors transfer to VLMs, and whether its use coincides with improved accuracy in a baseline computer vision task. 

The tendency of VLMs to use basic-level categories have significant utility in fields that depend on language models as agents. The findings of this paper suggest that some models prefer the most information-dense categorization layers for human communication, or the basic level. The recognition of this proclivity is useful for language model and human interactions because it helps us to understand the interactions between humans and VLMs. This paper also finds that the model's usage of this level of categorization coincides with the increased visual question answering accuracy that the literature finds in humans \citep{murphy_category_diff,rosch_basic_1976}.

To establish that VLMs use the basic level of categorization, we examine VLM categorization behavior to determine what alignment, if any, exists with human categorization via three experiments conducted on two recently released VLMs. The VLMs that were chosen were Llama 3.2 Vision Instruct (11B) and Molmo 7B-D for their popularity. Then to determine the benefits of this alignment with a specific focus on the impact of expertise prompting, an analysis of the relation between word properties and accuracy is performed. These forms of analysis gave us the following claims:

\begin{enumerate}[label=\arabic*.]
    \item Categorization Preferences

    \begin{enumerate}[label*=\arabic*.]
        \item VLMs tend to prefer human basic-level categories. We show this using a labeled set of images with basic level categories. This is important because it shows that there is overlap between the categorization behavior of VLMs and humans, in regards to the basic level.
        \item We observe a notable distinction in VLM basic-level categorization frequency between biological and non-biological objects, consistent with known human-categorization behavior. 
        \item VLMs exhibit a similar decrease in basic-level usage frequency under expert prompting to that shown in human experts. This was found by prompting the labeled basic category image dataset with an expert prompt. The shift in basic-level usage in human experts is associated with increased understanding of subordinate level categories. The increased understanding of subordinate level categories was not shown to exist in our testing, but the similarity in basic-level usage frequency justifies the need for investigation.
    \end{enumerate}
    \item Accuracy Analysis

    \begin{enumerate}[label*=\arabic*.]
        \item The basic prompt elicits basic-level usage, higher accuracy, and reduced responsiveness. The testing of accuracy and responsiveness in this claim refers to accuracy and tendency to respond with a possible answer on a visual question answering dataset. The tendency for the model to use the basic level was shown to exist without any special measures taken to enforce it. However, our observation also shows some slight utility in prompts that do not tend towards basic-level usage, as they could benefit from increased responsiveness at the cost of accuracy.     
        \item We do not observe an increased accuracy in expert prompted VLMs when they diverge from the basic-level. This claim is also determined by the analysis of the visual question answering dataset. The importance of this claim is that it provides a counter to the common belief that prompting for expertise is always better than not prompting for expertise. There already exists some evidence for this in the literature in regards to very small language models \citep{Multi_expert_Factuality}, but this claim provides further analysis of the distinction in utility that expert prompting provides.
    \end{enumerate}
\end{enumerate}

We show claims 1.1, 1.2, and 1.3 with an analysis of outputs generated by our model against a set of known basic level categories using different prompts and images. Claim 2.1 is shown through a testing of accuracy against a visual question answering task. Claim 2.2 is shown through an analysis of basic level usage compared to accuracy on a visual question answering task using word frequency as a proxy.

\section{Background and Related Work}

This section gives an overview of the elements of basic categorization that exist in the psychological literature and are examined in our experiments.

\subsection{Basic-Level Effect} 

The basic level is an emergent categorization level that appears in the many hierarchies humans use to distinguish objects from one another \citep{rosch_basic_1976}. This level is the most commonly used category when compared to both the less specific categories in our hierarchies (superordinate) and the more specific categories (subordinate). For example, when shown an image of a dog, a person will be more likely to describe it as a "dog" than an "animal" (superordinate) or by its specific breed (subordinate). One very commonly accepted explanation in the psychological field for why humans would have developed to use the basic level disproportionately, as opposed to a uniform categorization distribution, is that the basic categories have a higher information density and category differentiation \citep{expertice_basic,rosch_basic_1976}, permitting faster, more accurate, and more utilitarian categorization.

There are other recognized properties that occur when the basic level is invoked. These include being the most abstract level at which visual identification of stimuli can be aided by priming \citep{rosch_basic_1976}, children being able to sort basic objects taxonomically at ages when other visual stimuli would fail to be categorized \citep{rosch_basic_1976}, being the most frequently used categories in Common English \citep{rosch_basic_1976,ecoset}, and labels at this level being applied faster in visual stimuli \citep{rosch_basic_1976,expertice_basic}. These properties provide clear benefits to human cognition, and are reasonably explained by the higher information density and category differentiation the basic level exhibits. 

\subsubsection{Biological vs. Non-Biological}

The basic level also has a notable distinction in the cognitive science literature between biological and non-biological objects. The distinction is that biological things seem to be associated with each other based on their features, such as fangs, claws, or legs, while non-biological objects tend to be associated with each other based on attributes associated with function \citep{rosch_basic_1976}. The strength of this difference is hard to quantify due to the different ways in which the distinction between biological and non-biological things can be defined. However, existing evidence has indicated that non-biological objects have a higher tendency to be described with a subordinate category than the biological things \citep{rosch_basic_1976}. 

\subsubsection{Expert basic-level Shift}

The cognitive science literature also suggests that expertise changes the expression of basic level effects. It has been shown that the highest information density for an expert tends to be below the typical basic level for subject matter non-experts \citep{expertice_basic}. This suggests that the basic level has shifted lower to match the more granular understanding experts have within their domain. For the sake of clarity, when referring to this subordinate shifted basic level, it is called the expert-shifted-basic level, and any usage of the phrase "basic level" not specified refers to the basic level observed in non-experts. 

\subsection{LLM Cognitive Categorization Behaviors}

Many studies have investigated human-like cognitive effects in LLMs. Of particular relevance are those that study human-like behaviors related to categorization in language models \citep{misra2021language, roberts2024using, vemuri2024well,roberts2024large}.  The existing works examine patterns in human categorization such as the fan and typicality effects, and their applicability to LLM categorization. Papers on the typicality effect generally find that typicality effects present themselves in LLMs, but tend to present themselves more strongly in combined vision and text input models than in only vision or only text based models \citep{misra2021language,vemuri2024well}. The literature on the fan effect demonstrates that the recall uncertainty for an object demonstrated by a model is in part accounted for by the fan effect \citep{roberts2024large}. 

The fact that these cognitive effects, first established in human categorization, have been found in LLMs promises other effects are likely present as well. Despite this, no existing literature for the basic level could be found regarding LLMs or VLMs. This was the largest motivating factor for the efforts presented in this paper.

\section{Basic-level Preferences}
This section investigates whether the tested VLMs share the human preference for basic-level categories. It considers distinctions found in the literature, such as the distinction between expert and non-expert usage, and between biological and non-biological categories. Using a large dataset of images paired with known basic-level categories, we prompt our models with expert and non-expert instructions to label each image and compare their outputs to the ground truth labels. We then analyze these results through a comparison to the known human categorization behavior to assess the degree of alignment with the tested models.

\subsection{Ecoset Dataset}
The dataset of images used for categorization comes from an organization called KietzmannLab, specifically their Ecoset dataset \citep{ecoset}. This is a dataset of about 1.5 million images all falling into one of 565 basic categories, with the basic categories determined based on frequency of use.

The categorizations were manually curated by humans, and involved removing the less "concrete" categories. For this paper, we used a subset of 28,250 images total, with 50 images from e ach category.

It is worth noting that the basic-level image dataset contains some noisy labels which are likely due to the curation process. A sample of 530 images was checked and only one categorization error was found. However, it should still be noted that these errors exist in the dataset.

\subsection{Prompt Design}

\begin{figure}[!h]
    \centering
    \resizebox{0.8\linewidth}{!}{ 
    \begin{tcolorbox}[expertstyle]
        \hfill \textbf{Human} \\
        Describe the main subject in this image in minimal words

        [IMAGE]
    \end{tcolorbox}
    } 

    \vspace{2mm} 

    \resizebox{0.8\linewidth}{!}{ 
    \begin{tcolorbox}[llmstyle]
        \textbf{VLM} \\

        [RESPONSE]

    \end{tcolorbox}
    } 
    
    \caption{Base prompt to elicit VLM description. The VLM response is then compared to the identified basic-level description.}
    \label{fig:Prompt1}

\end{figure}

\begin{figure}[!h]
    \centering
    \resizebox{0.8\linewidth}{!}{ 
    \begin{tcolorbox}[expertstyle]
        \hfill \textbf{Human} \\
        ~``Say the word for experts in the field of whatever the primary object in the image is, then say that you are one of whatever the term is. Then acting as this expert, describe the main subject in minimal words''

        [IMAGE]
    \end{tcolorbox}
    } 

    \vspace{2mm} 

    \resizebox{0.8\linewidth}{!}{ 
    \begin{tcolorbox}[llmstyle]
        \textbf{VLM} \\

        [RESPONSE]

    \end{tcolorbox}
    } 
    
    \caption{Expert prompt to elicit VLM description. The VLM response is then compared to the identified basic-level description.}
    \label{fig:Prompt2}

\end{figure}
The prompt seen in Figure \ref{fig:Prompt1} is used as a baseline for comparison to the expert prompting style. It asks a model to use minimal words as a method of preventing incidental use of basic-level categories as a result of verbose responses. This is meant to mitigate instances in which an individual output contains identifiers from multiple categorization levels.
This prevents a misidentification of a preference for the basic level as a result of long answers.

The prompt seen in Figure \ref{fig:Prompt2} is our expert prompt. It is constructed such that it follows the guidance of \citet{xu_expertprompting_2025} as best as possible. The predicted behavior of this prompting style based on the cognitive literature is a decrease in the tendency to use the basic-level

\subsection{Two Proportion Z-testing}
We consider the presence of a basic level description to be the result of a Bernoulli trial and examine the relative rate of success. To test the statistical significance of the shift in the measured proportions, a two proportion Z-test is employed. The two proportion Z-test can be used to determine the level of confidence that a sampled proportion differs between populations. This measure is used to evaluate the statistical significance of the difference in basic level usage as a proportion between both the biological and non-biological samples, and the non-expert and expert samples.

The assumptions of the Z-test are that the test statistic, the average number of successes in the trial, is normally distributed. This is an accurate assumption in a large sample due to the central value theorem for Bernoulli trials.

\subsection{Basic Level Usage: Claim 1.1}

\begin{table}[!h]
    \centering
    \caption{Llama 3.2 and Molmo both show basic level usage rates significantly higher than random selection over superordinate, basic, and subordinate levels.}
    \begin{tabular}{|c|c|c|c|}
    \hline
    {\bf Model}&{\bf Sample size}  & \bf {Basic level} & {\bf Basic Rate} \\
    \hline
    {Llama}         & {28,250} & {17,030} & {60\%} \\
    \hline
    {Molmo}         &{28,250}    & {14,766} &  {52\%} \\
    \hline
    \end{tabular}
    \label{tab:Basic Categorization}
\end{table}

The initial prompt results in 17,030 outputs containing the basic-level category in 28,250 total runs across the set of categories in the dataset, resulting in a 60.2\% basic categorization in the outputs for the Llama model. There is a notable decrease in the Molmo model compared to the Llama model, with 14,766 total outputs using the basic-level categorizations, or 52.269\% of all outputs using the basic category. However, both models are above 50\% of outputs containing the human basic-level. Considering the preference for the basic level in humans is a simple plurality, a result of over 50\% of our outputs containing the basic level indicates that our model has a majority preference for the basic level. This indicates that our models have at least as strong as a preference for the basic level as is seen in humans.


\subsection{Biological and Non-biological Categories: Claim 1.2}

Next we examine the basic-level categorization impacts between biological and non-biological objects. The biological categories include 204 of the 565 total categories, with the non-biological categories being the remaining 361. The results show a notable drop of 5.3\% for the Llama model from the base prompt biological versus non-biological testing. There is also a respective drop in the Molmo Testing of 3.4\%. While the value change is low, the results are highly significant due to the large sample size used in evaluation. 


\begin{table}[]
    \caption{VLMs use the basic level significantly less (p<0.01) for non-biological items}
    \centering
    \begin{tabular}{rlc}
     & \textbf{Llama 3.2} & \textbf{Molmo} \\
    \multicolumn{1}{r}{} & \multicolumn{2}{c}{Basic Level Usage} Total (\%) \\
    \toprule
    \textbf{Base Biological} & 6,425 (64\%) & 5,507 (55\%) \\
    \textbf{Base Non-Biological} & 10,605 (58\%) & 9,259 (51\%) \\
    \bottomrule
    \textbf{Z-test} & 38 (p<0.01) & 30 (p<0.01) 
    \end{tabular}%
    \label{tab:biovnonbio}
\end{table}

The results in Table \ref{tab:biovnonbio} show the p-value is less than 0.01 for both model test statistics. This strongly suggests that there is a notable difference in the distributions of basic level usage between biological and non-biological categories in both tested models, consistent with the difference observed in the cognitive literature regarding human categorization behavior.

\subsection{Expertise prompting: Claim 1.3}

We found an approximately 5.8 percentage point drop from the percentage of basic-level categorization that occurred from the initial non-expert prompt in the Llama model and 3 percentage point drop in the Molmo model.
We also note that each of the differences are statistically significant (p < .01). These results can be seen in Table \ref{tab:exp_total}.

\begin{table}[!h]
    \caption{VLMs use the basic-level significantly less (p<0.01) when prompted to act as an expert.}
    \label{tab:exp_total}
    \centering
    \begin{tabular}{rlc}
     & \textbf{Llama 3.2} & \textbf{Molmo} \\
    \multicolumn{1}{r}{} & \multicolumn{2}{c}{Basic Level Usage} Total (\%) \\
    \toprule
    \textbf{Base} & 17,030 (60\%) & 14,766 (52\%) \\
    \textbf{Expert} & 15,380 (54\%) & 13,901 (49\%) \\
    \bottomrule
    \textbf{Z-test} & 77 (p<0.01) & 17 (p<0.01) 
    \end{tabular}%
\end{table}

We also note an interesting relationship between expert and non-expert prompting and biological vs non-biological categories. The differences in basic category usage between biological and non-biological categories expand when expert prompting is applied to the Llama model, but the differences shrink when expert prompting is applied to the Molmo model.

\begin{table}[]
    \caption{VLMs use the basic-level significantly less (p<0.01) for non-biological categories.}
    \label{tab:expbiononbio}
    \centering
    \begin{tabular}{rlc}
     & \textbf{Llama 3.2} & \textbf{Molmo} \\
    \multicolumn{1}{r}{} & \multicolumn{2}{c}{Basic Level Usage} Total (\%) \\
    \toprule
    \textbf{Expert Biological} & 6,120 (61\%) & 4,992 (49.4\%) \\
    \textbf{Expert Non-Biological} & 9,260 (51\%) & 8,909 (49.1\%) \\
    \bottomrule
    \textbf{Z-test} & 50 (p<0.01) & 9 (p<0.01) 
    \end{tabular}%
\end{table}

\section{The Basic Level and VQA accuracy}

This section examines the relationship between the use of the basic-level and accuracy on visual question answering (VQA) tasks. This relationship is examined in relation to a basic prompt that was shown in the previous section to invoke the basic level and to two prompts which request the model to respond as an expert. The primary methodology for this section is an analysis of word frequency and VQA accuracy using the Llama 3.2 Model. This section first uses word-frequency analysis to show the presence of basic-level usage tendencies in outputs to the VQA task. Then we evaluate accuracy on the VQA test. Finally combine these two metrics to determine how word frequency, our indicator of basic-level categorization, relates to task accuracy over the different prompt types.

\subsection{Dataset Description}
The dataset used to determine accuracy was the version one multiple choice visual question answering (VQA) dataset created by Virginia Tech \citep{VQA}. Specifically, the validation split consisting of 121,512 multiple choice questions with 1,215,120 human-generated answers was used. These questions and answers span a total of 40,504 images, with each question having 10 human generated answers, and belonging to one of 65 question categories. However, due to computational limitations, not all 121,512 questions were used.

In accordance with the accuracy measurements of \citet{VQA}, the accuracy for this dataset is determined by comparing the number of times a given answer from the tested VLM was also given by the 10 human respondents. This number is then divided by three and run through the standard min function compared to 1. The resulting number is multiplied by 100 to give an accuracy percentage. Dividing by 3 accounts for variability among human answers and for answers that were not agreed upon as a consensus, but were still reasonable.

This can be expressed as the formula in Equation \ref{fig:accuracy_equation}, where N is the number of human respondents that gave the same answer as the model.

\begin{equation}
    Accuracy = 100 * min(N/3,1)
\label{fig:accuracy_equation}
\end{equation}

For example, if our model is asked a question and responds with ~``yes'' and 2 of the 10 human responses also said yes, then the accuracy would be 66\%. This style of accuracy measurement was generated by the creators of the dataset and is not unique to this paper.

\subsection{Prompting Styles}

\tcbset{
    mymsg/.style={
        enhanced,
        colframe=white, frame hidden,
        boxrule=0pt,
        arc=18pt,
        outer arc=18pt,
        boxsep=2pt, left=8pt, right=8pt, top=6pt,
        colback=#1!25,
        fuzzy shadow={1mm}{-0.7mm}{0mm}{0.1mm}{black!40!white},
        width=0.8\linewidth,
        before skip=-0.5mm, 
    },
    initialstyle/.style={mymsg=green, enlarge right by=10mm},
    dynamicstyle/.style={mymsg=red, enlarge right by=10mm},
    staticstyle/.style={mymsg=blue, enlarge right by=10mm},
    fakestyle/.style={mymsg=blue, enlarge right by=10mm},
    responsestyle/.style={mymsg=blue, enlarge right by=10mm}
    
}

\begin{figure}[!h]
    \centering
    \resizebox{0.8\linewidth}{!}{ 
    \begin{tcolorbox}[initialstyle]
        \hfill \textbf{Basic Prompt} \\
        ~``Answer the following question using one of the given choices. '' + question + ~``The choices are ''+ multiple choices

        [IMAGE]
    \end{tcolorbox}
    } 

    \resizebox{0.8\linewidth}{!}{ 
    \begin{tcolorbox}[staticstyle]
        \hfill \textbf{Simple Expert} \\
        ~``Imagining you are an expert in the regarding field, try to answer the following instruction as professional as possible.'' + question + ~``The choices are ''+ multiple choices

        [IMAGE]
    \end{tcolorbox}
    } 
    \resizebox{0.8\linewidth}{!}{ 
    \begin{tcolorbox}[dynamicstyle]
        \hfill \textbf{Advanced-Expert Prompt} \\
        ~``Say the word for experts in the field of whatever the primary object in the image is, then say that you are one of whatever the term is. Then acting as an this expert, answer the following multiple choice question. '' + question + ~``The choices are ''+ multiple choices

        [IMAGE]
    \end{tcolorbox}
    } 

    \vspace{2mm} 

    \caption{The Structure of The Three Prompts Tested in This Section}
    \label{fig:PromptAccuracy}

\end{figure}

Three prompting styles are tested: one to elicit the basic level and two to simulate expertise. These are shown in Figure \ref{fig:PromptAccuracy}.

The basic prompt and Advanced-Expert prompt mimic the structures of the previous initial and expert prompt with the new question type replacing the old one. We also added an additional Simple Expert prompting style, which follows the static prompting method seen in \citet{xu_expertprompting_2025}. This was done as the Simple Expert is very similar to how expert prompting is often done. This gives our results slightly more comparability to general expert prompting usage.

\subsection{Output Generation and Processing}
\label{sec:Output Gen}
The results of this section were created by allowing the models to generate outputs over the prompts, along with their associated images, for approximately 15 hours for each prompt. The max token count was set to thirty for both the basic and Simple-Expert counts, and 60 for the Advanced-Expert prompt, because of the additional tokens the Advanced-Expert prompt needed to complete its first instruction. 

There was a large difference in the total number of outputs generated by each prompt, so steps were taken to avoid biasing the analysis of the outputs.
Due to time constraints and increased computational load, only 30600 contexts from the original dataset were used for the Advanced-Expert prompting experimental subset. We were concerned that the difficulty or implicit word frequency of the correct answers would be different between the sampled 30,600 and the higher total number that were completed by the other two prompts. To avoid this the basic and Simple-Expert prompting style both had their outputs cut down to only the 30,600 responses corresponding to questions the Advanced-Expert prompt completed. Except with the specific sections that analyze total samples, every analysis in this section was performed using the outputs from these 30,600 questions.

Outputs were analyzed by matching model responses to multiple-choice answers. For example we could want to prompt our model with a basic prompting method for a question related to an image of a couch. We can see a real example from the VQA dataset of this exchange and the generated response in Figure \ref{fig:PromptAccuracy}.

\begin{figure}[!h]
    \centering
    \resizebox{0.8\linewidth}{!}{ 
    \begin{tcolorbox}[initialstyle]
        \hfill \textbf{Model Input} \\
        ~``Answer the following question using one of the given choices. What color is the couch? The Choices are  3, orange, no, dane101, 2, 1, brown, potatoes, 4, yes, red, 1 ton, white, black, scrapbook, green, horse race, blue ''
        [IMAGE]
    \end{tcolorbox}
    } 
    \resizebox{0.8\linewidth}{!}{ 
    \begin{tcolorbox}[llmstyle]
        \hfill \textbf{Model Response} \\
        "The couch is green"
    \end{tcolorbox}
    } 
    \caption{Real VQA Prompt and Response}
    \label{fig:PromptResponse}
\end{figure}

This exchange would be evaluated using just the word ~``green'' using the accuracy equation from Equation \ref{fig:accuracy_equation}. This type of analysis would result in non-responsive answers, or answers that did not include one of the multiple choices, getting an accuracy of 0\%.  This could result in accuracy figures that were biased toward answers aligning with human responses, instead of factual accuracy, which negatively impacts our ability to test our hypothesis. However, this potential issue is due to the pre-established method of measuring accuracy the VQA dataset creators decided on, and our decision against the use of a testing method called \textit{cloze} testing.

To mitigate this, we decide to identify two classes of outputs, ~``responsive'' and ~``non-responsive''. These are outputs that either had a VQA multiple choice answer keyword, or outputs that did not, respectively. The differences in responsiveness between the prompt types is noted and documented in our results.

\subsection{A Note On Cloze Testing}
We mentioned in Section~\ref{sec:Output Gen} a type of analysis called \textit{cloze testing}. This is a common form of analysis that is applied to language models when responding to multiple choice tests. In this form of testing, the model's accuracy is evaluated based on its ability to complete fill-in-the-blank prompts. For three main reasons, it was not suitable for this paper.

First, there is the methodological limitation that comes with the repeat back instruction in the Advanced-Expert prompting. This limits our ability to generically determine where the specific word or phrase that would most appropriately fill in the blank is in the output. This reason by itself makes cloze testing unsuitable for this paper, as it functionally removes our ability to use the Advanced-Expert prompt style.

Second, cloze analysis does not align with how models are typically used. Most language model usage is defined by a user typing in a prompt and getting an unrestricted and full output. This is distinct from being limited to fill-in-the-blank style tasks. Our accuracy testing is more representative of practical model usage.

Third, a cloze analysis would limit the ability of the model to generate robust outputs that would place the basic-level categories in context. Our chosen method of analysis lets a model produce a longer and more natural response.

\subsection{Basic-Level Usage Consistency}

We demonstrated in the previous section that the tested models used the basic level preferentially. We perform additional analysis to support the presence of this preference in the new task type. To evaluate this, average word frequency was compared between the basic and expert prompts for both the Ecoset and VQA tests. This metric has been shown in the literature to serve as a proxy for basic-level usage \citep{average_word,rosch_basic_1976}. The utility of this comparison is that it helps assess whether the difference in basic-level usage between basic and expert prompting is consistent from the Ecoset task to the VQA task. The specific goal was to determine whether the basic prompt continues to elicit basic-level categories more frequently than the expert prompt.

\subsection{Word Frequency Consistency}

Word frequency was measured by computing the common-use frequency of each word in the generated outputs using the \textit{'wordfreq'} python library and averaging those values across all words in each output. These responses were then grouped into \textit{"responsive"} and \textit{"non-responsive"} as well as question type.

Multiple stages of cleaning were performed, the first of which was the removal of stop words.
Stop words are commonly recognized sets of sentence shaping words that would add noise to the frequency of the categorization. They are words such as \textit{~``The'',~``i'',~``and'',~``between'',~``than''}, and \textit{~``should'}. There were removed for our first set of testing.

Before we  present these results, it is important to clarify what the cognitive literature predicts. The cognitive literature notes that basic categories have a much higher use frequency than subordinate or superordinate categories \citep{rosch_basic_1976}. This is one of the defining features of basic categories. If we combine this with our earlier finding that the expert prompt invokes basic-level categories less often than the basic prompt, we should expect lower word frequency from the expert-prompted outputs. 

This behavior was confirmed in both responsive and non-responsive VQA outputs after stop word removal. However, we also needed to account for the words which were repeated from the prompt itself, or that were apparent restatements of the prompt in the response. We examined the Ecoset dataset first and found a set of words that were repeated and clearly came from the prompts. To adjust for this issue, we remove the words \textit{~``main'',~``expert'',~``subject'',} and \textit{~``one''}. 

\begin{table}[]
    \caption{Word Frequency with Stop Words and Prompt Repeated Words Removed, Responsive Only}
    \centering
    \begin{tabular}{c|l|c}
         Test Set & Prompt & Average Word Frequency \\
        \toprule
        \multirow{2}{*}{Ecoset} & 
        Initial & 0.00009339 \\ 
        & Expert & 0.00007007 \\
        \midrule
        \multirow{2}{*}{VQA Responsive} & Basic & 0.00030843 \\
        & Advanced-Expert & 0.00016984
    \end{tabular}
    \label{tab:VQA_word_freq_IED}
\end{table}

A similar analysis was performed on the VQA Advanced-Expert outputs. The commonly repeated words associated with the prompts were \textit{~``expert'',~``field'',~``primary'',~``one'',~``main'',~``answer''}, and \textit{~``image''}. After accounting for these repeated words and stopwords, the final frequency results are shown in Table \ref{tab:VQA_word_freq_IED}. These adjusted results align with expectations from the cognitive literature.

\subsection{Accuracy: Claim 2.1}

\begin{table}[!h]
    \caption{Accuracy by prompt type and responsiveness}
    \label{tab:Accuracy_all}
    \centering
    \begin{tabular}{c|c|c|c|c}
         Prompt &Total Samples & \% Non-Responsive & Responsive Accuracy & Total Accuracy \\
        \hline
        Advanced-Expert & 30,600 & 16.7 & 55.08 & 45.91\\
        Simple-Expert & 49,800 & 24 & 56.43 & 42.76\\
        Basic & 74,600& 22.7 & 68.49 & 56.90
    \end{tabular}
    
\end{table}
\begin{table}[!h]
    \caption{Accuracy of the Basic and Simple-Expert prompting After Question Matching To Total Advanced Questions}
    \centering
    \begin{tabular}{c|c|c|c}
         Prompt & Responsive Only & Total Samples & Accuracy \\
        \hline
        Simple-Expert & Responsive Only & 22,979 & 56.48 \\
        Simple-Expert & Non-Responsive Included & 30,600 & 42.48\\
        Basic & Responsive Only & 23,598 & 68.47 \\
        Basic & Non-Responsive Included & 30,600 & 56.93
    \end{tabular}
    
    \label{tab:accuracy_shrunk}
\end{table}
The first analysis disregarded question categories and measured overall accuracy across both responsive and non-responsive outputs. These results can be seen in Table \ref{tab:Accuracy_all}.

Additionally, accuracy results after matching the basic and Simple-Expert questions to the Advanced-Expert question set can be seen in Table \ref{tab:accuracy_shrunk}. This adjustment had minimal impact on accuracy. It slightly increased responsiveness in the basic prompt compared to the Simple-Expert, but the difference was small to begin with and thus not considered meaningful.

These results run contrary to the initial hypothesis that the two expert prompting styles would be more accurate.

A notable trend is the higher responsiveness observed in Advanced-Expert outputs, which appears consistent with findings from \citet{xu_expertprompting_2025} in its testing on preference ranking. This was not explored in detail, but may indicate that our assumption of the image fulfilling the “informative” criterion in expert prompting was effective. We believe this increases the applicability of our results.

\subsection{Accuracy by Category}

\begin{figure}[h]
    \centering
    \includegraphics[width=\linewidth]{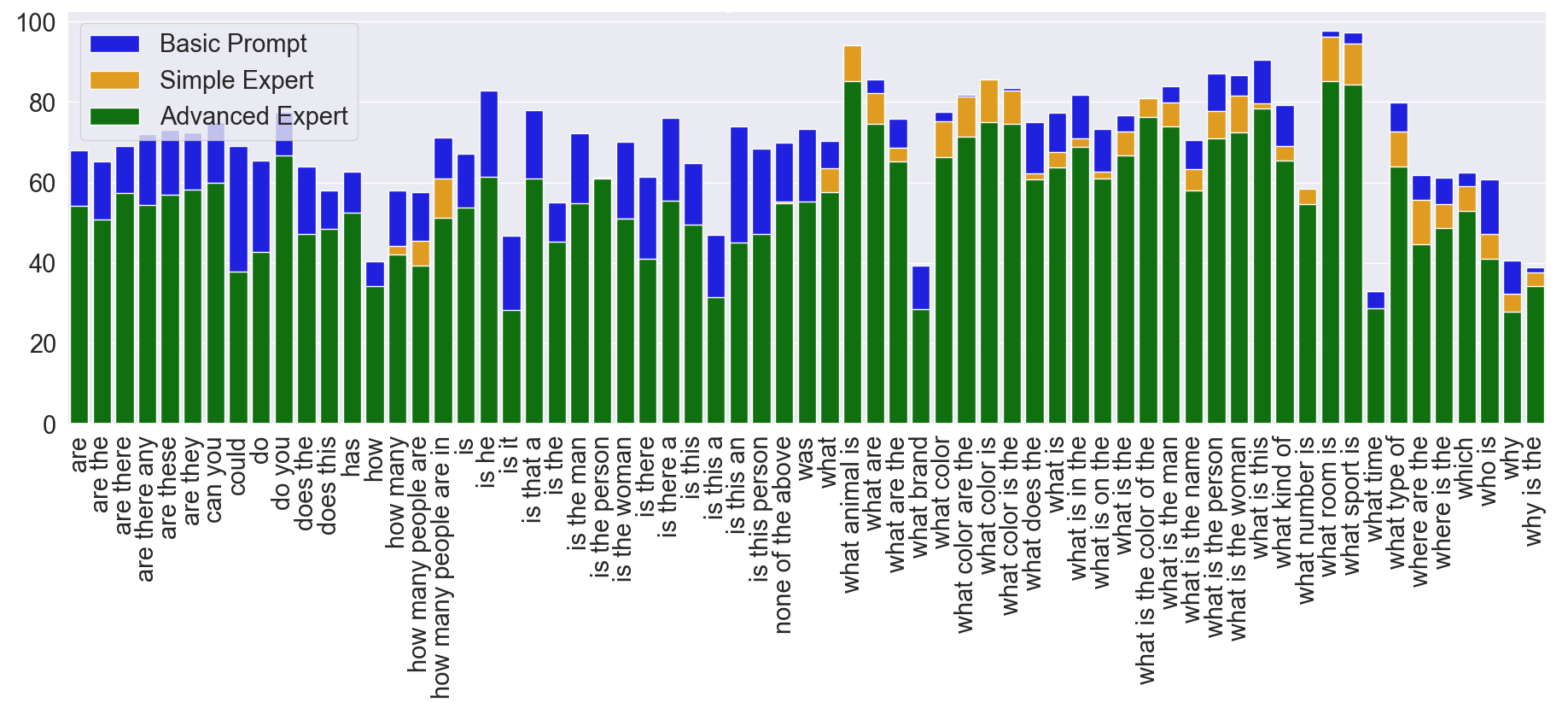}
    \caption{The Most Accurate (y-axis) Prompting Style Is Basic for the Majority of the Question Types (x-axis)}
    \label{fig:Responsive Accuracy}
\end{figure}

The category-specific analysis of the VQA dataset's question categories reveals a trend consistent with the overall accuracy results. The basic prompting tends to be more accurate than both the simple and Advanced-Expert prompting. These results are laid out in Figure \ref{fig:Responsive Accuracy}. 

While the accuracy differences are small across many question types, the basic prompt generally produces the highest accuracy. The simple and Advanced-Expert prompts typically compete for second place.

\subsection{Word Frequency and Accuracy}

The remaining subsections will focus on the relationships between word frequency and accuracy. It is important to clarify why we are examining these two factors together.

Accuracy on VQA tasks is something that we would expect to improve with object recognition abilities. We also separately know that object recognition is improved in humans through the use of the basic level \citep{rosch_basic_1976,johnson1997effects}. This means that if we observe an increase in basic-level usage in our models, we would expect to see a corresponding increase in VQA accuracy if the model benefits from the basic level similar to how humans do.

We use word frequency as a proxy for basic-level categorization because the cognitive literature consistently demonstrates that basic-level terms appear more frequently in common use \citep{Fisher_1987,expertice_basic,johnson1997effects}. By analyzing correlations between word frequency patterns and VQA accuracy, we can evaluate whether increased-basic level usage contributes to improved task performance. We can combine this with the accuracy results across prompts so that we can find evidence for or against the existence of the utility the basic level provides each prompt type in terms of VQA accuracy.

\subsection{Analysis Methodology}

The methodology begins by grouping outputs by the question category used to generate them. These were the question type labels that the VQA dataset had associated with each question. This controls for potential bias introduced by categories with disproportionate sample sizes.

Then we measure the average word frequency and accuracy by question type. The python library we used to calculate word frequency draws from a large number of sources of natural language use. It calculates frequency by dividing the number of times a unique word appeared in its corpus by the total number of words in its amalgamated sources of text.

For this paper, outputs containing basic-level terms are expected to yield average word frequencies closer to 1.  For example, the word “the,” the most common word in English, appears in roughly 1 out of every 15 to 20 words and is assigned a value of 0.0537 by wordfreq. To reduce noise in our data, stop words were removed, consistent with earlier basic-level usage analysis.

Pearson correlation coefficients between accuracy and average word frequency were calculated for each prompting style within each question type. The results for this are presented in Table \ref{tab:Accuracy_vs_word_freq_stats_combined}. Associated p-values test for each r-value are seen next to their relevant r-values.

\subsection{Correlation Analysis: Claim 2.2}

\begin{figure}
    \centering
    \includegraphics[width=0.75\linewidth]{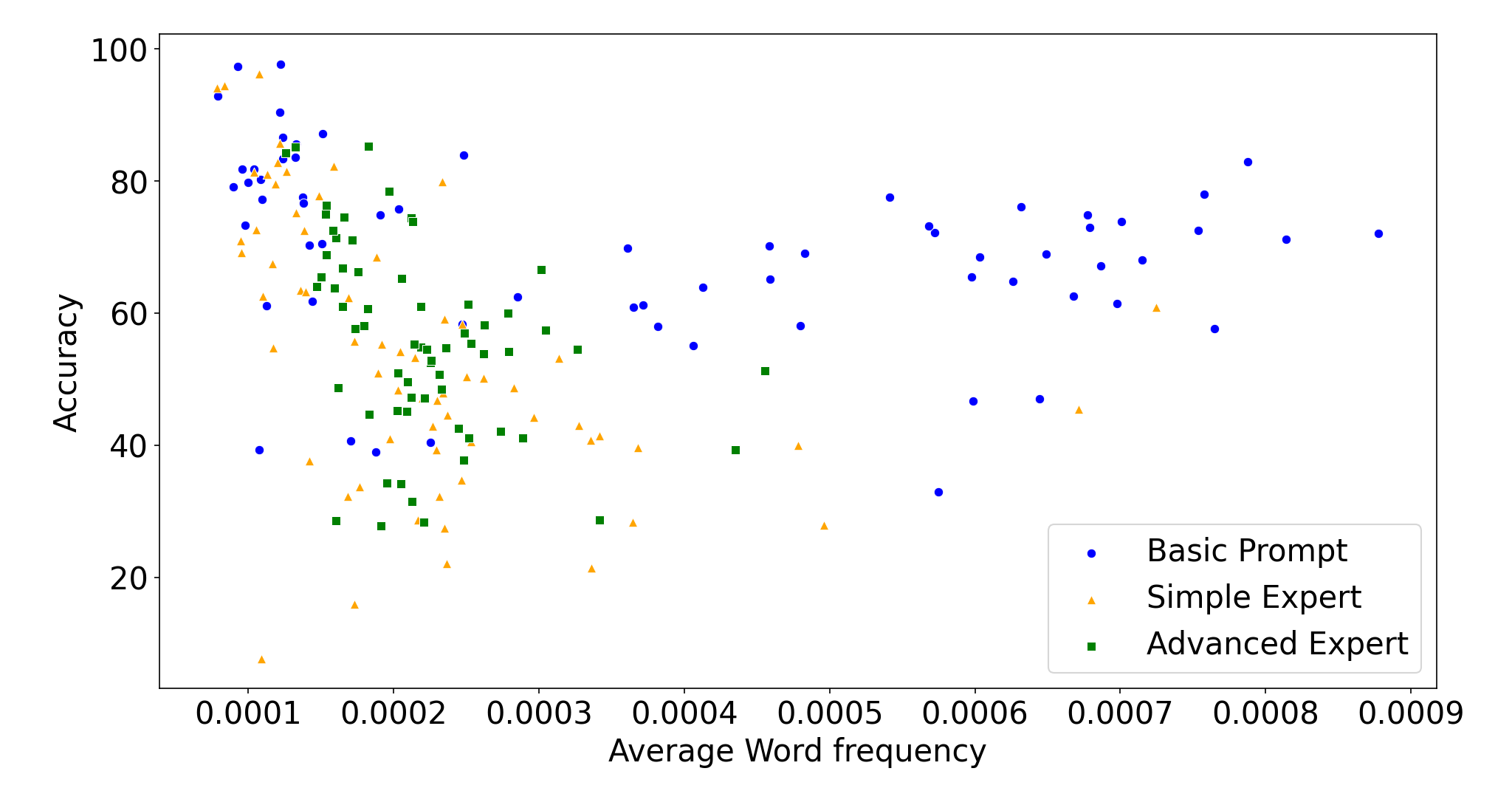}
    \caption{Scatter Plot of Accuracy vs Word Frequency by Category With Stop Words Removed, Responsive.}
    \label{fig:accuracy_vs_word_freq}
    
\end{figure}

The word frequency results for the basic, Advanced Expert and Simple-Expert prompts are shown in Figure \ref{fig:accuracy_vs_word_freq}, with each point representing a unique question type. The relevant correlation analysis is displayed in Table \ref{tab:Accuracy_vs_word_freq_stats_combined}.

\begin{table}[]
    \caption{Accuracy vs Word Frequency By Category With Stop words Removed Correlations, Responsive}
    \label{tab:Accuracy_vs_word_freq_stats_combined}
    \centering
    \begin{tabular}{c|c|c}
         Prompt & r-value & p-value \\
        \toprule
        Basic & -.25 & .0.018154 \\
        Simple Expert & -.42 & 0.000422\\
        Advanced Expert & -.40 & 0.000884\\
    \end{tabular}
    
\end{table}

The results show a very weak negative correlation between word frequency and accuracy for the basic prompt, and stronger negative correlations for both expert prompts. This outcome aligns with our predictions. The basic prompt’s correlation coefficient is below the common threshold of |0.3| in magnitude, beyond which variables are typically considered weakly related. This would put the basic-level prompts in an area of correlation that indicates negligible or no correlation. These results can be seen in Table \ref{tab:Accuracy_vs_word_freq_stats_combined}. 

\section{Conclusions}
Our experiments demonstrate the propensity of VLMs towards basic categories, consistent with the distinctions observed in humans. We also observe an improved accuracy in VQA tasks using our basic prompt, but with a general decrease in accuracy in our expert prompting methods associated with the basic level. This is a surprising departure from the common perception that expertise prompting typically improves task performance.

The findings in this paper should serve as a potential warning against over reliance on expertise prompting. We find that expertise prompting can fail to result in more accurate outputs, and we also find that this tendency correlates negatively with the preference of the expertise prompting methods to use the basic level. There is room in the future to explore the applicability of these results on larger models, and on how models or prompting styles that show positive results for expertise prompting interact with the basic level.

We believe that aligning language model usage more closely with human cognition will fundamentally improve their abilities. By exploring the middle ground between human and language model cognition, this work will hopefully be a contributive tool for the growing domain of language model cognition.





{\parindent -10pt\leftskip 10pt\noindent
\bibliographystyle{cogsysapa}
\bibliography{format}

}


\end{document}